\documentclass[journal]{journal}
\usepackage{caption}

\usepackage{times}

\usepackage{soul}
\usepackage{url}
\usepackage[hidelinks]{hyperref}
\usepackage[utf8]{inputenc}
\usepackage{multicol}
\usepackage{amsmath}
\usepackage{float}
\usepackage{graphicx}
\usepackage{amsmath}
\usepackage{booktabs}
\usepackage{enumerate}
\usepackage{times}
\usepackage{booktabs}
\usepackage{longtable}
\usepackage[noend]{algpseudocode}
\usepackage{fancyhdr,graphicx,amsmath,amssymb}
\usepackage{afterpage}
\usepackage{times}

\usepackage{soul}
\usepackage{url}
\usepackage[hidelinks]{hyperref}
\usepackage[utf8]{inputenc}
\usepackage{graphicx}
\usepackage{amsmath}
\usepackage{booktabs}
\makeatletter
\renewcommand{\Function}[2]{%
  \csname ALG@cmd@\ALG@L @Function\endcsname{#1}{#2}%
  \def\jayden@currentfunction{#1}%
}
\newcommand{\funclabel}[1]{%
  \@bsphack
  \protected@write\@auxout{}{%
    \string\newlabel{#1}{{\jayden@currentfunction}{\thepage}}%
  }%
  \@esphack
}
\makeatother

\usepackage[ruled,vlined]{algorithm2e}
\include{pythonlisting}
\usepackage{lipsum}
\usepackage[
backend=biber,
style=ieee,
]{biblatex}
\usepackage{svg}
\ifCLASSINFOpdf
\else
\fi
\begin{document}
%
\title{MULTI-FLGANs: Multi-Distributed Adversarial Networks for Non-IID distribution}
%
%
%

\author{Akash~Amalan,
        Rui~Wang,
        Yanqi~Qiao,
        Emmanouil~Panaousis,~\IEEEmembership{Member,~IEEE,}
        and~Kaitai~Liang,~\IEEEmembership{Member,~IEEE}
\thanks{A. Amalan, R. Wang, Y. Qiao and K. Liang are with the Department
of Intelligent Systems, Delft University of Technology, 2628 XE, Delft, the Netherlands.}
\thanks{E. Panaousis is with the School of Computing and
Mathematical Sciences, University of Greenwich, SE10 9LS London, UK.
E-mail: e.panaousis@greenwich.ac.uk.}
\thanks{Manuscript received April 19, 2005; revised January 11, 2007.}}

%
%

\markboth{Journal of \LaTeX\ Class Files,~Vol.~6, No.~1, January~2007}%
{Shell \MakeLowercase{\textit{et al.}}: Bare Demo of IEEEtran.cls for Journals}
%



\maketitle
\thispagestyle{empty}

\begin{abstract}
Federated learning is an emerging concept in the domain of distributed machine learning. This concept has enabled GANs to benefit from the rich distributed training data while preserving privacy.  However, in a non-iid setting, current federated GAN architectures are  unstable, struggling to learn the distinct features and  vulnerable to mode collapse. In this paper, we propose a novel architecture MULTI-FLGAN to solve the problem of low-quality images, mode collapse and instability for non-iid datasets. Our results show that MULTI-FLGAN is four times as stable and performant (i.e. high inception score) on average over 20 clients compared to baseline FLGAN.

\end{abstract}


%
\IEEEpeerreviewmaketitle

\section{Introduction}
\IEEEPARstart{G}{eneral} Adversarial Networks (GANs)~\cite{brownlee_2019, Isola_2017_CVPR} have been used in applications ranging from style transfer to image-to-image translation. While they are widely used, they suffer from three main problems:
\begin{itemize}
    \item A lack of data affects their ability to generate high-quality images~\cite{NVIDA,DiffAugmentationGAN}.
    \item They are vulnerable to mode collapse, a case in which a GAN only generates a specific subset of the real images~\cite{mao_li_2020, DBLP:journals/corr/abs-1806-11382}. 
    \item They can be unstable (i.e. fluctuations in IS score), failing to converge or visibly improve~\cite{DBLP:journals/corr/abs-1806-11382}. 
\end{itemize}

Several architectures~\cite{Cycle_GAN, NVIDA, DiffAugmentationGAN, DataEffecient-GAN} were proposed to address the first problem through traditional and differential data augmentation methods. Federated learning (FL), an emerging concept~\cite{FL}, also showed promise in overcoming the inability of GANs to generate high-quality images when lacking data. This paper will address this problem by adapting GANs to a distributed setting through FL. In FL, each client uses their private datasets to jointly learn the global model. The clients trains the model locally with their training data and send the updated weights to the central server to aggregate and update the global model. Thus, FL benefits from rich distributed training data while also potentially preserving the client’s privacy.

Similarly, many architectures were proposed to address the last two problems. For instance, Ishan et al.  \cite{GMAN} proposed GMAN, a generic architecture with one generator and multiple discriminators. By introducing a one vs all game where the generator tries to fool many discriminators, the chance of mode collapse is significantly reduced. Likewise, Q. Hoang et al.~\cite{MGAN} suggested a Multi-Generator GAN (MGAN) with an arbitrary grouping of generators and discriminators to mitigate instability and improve convergence. 

FLGAN (or FedAvgGan)~\cite{FLGAN} was the first architecture to bridge the gap between FL and GANs by assigning a generator and discriminator to each client. For each iteration, FLGAN uses the aggregated weights~\cite{fedAvg} of local models to update the global model, i.e. global generator and discriminator. However, this baseline architecture was found to be vulnerable to mode collapse~\cite{FedAVGDis} and privacy leaks~\cite{privacyattacks}. 

Although other variants~\cite{GSWGAN, DPGAN} were proposed to improve the performance and privacy of the FLGAN architecture, Hardy et al.~\cite{Hardy_2019}. were the first to suggest decreasing the computation cost at worker nodes. They did this by introducing MDGAN, a novel architecture that extends GMAN to a distributed setting with a central generator at the server and a discriminator per client. To elaborate, MDGAN swaps each discriminator every \textit{k} iteration and aggregates their weights using fedAvg. This way, a one vs all game is introduced where the generator at the server attempts to fool all discriminators in each client node. By decreasing the computation cost for each client node, MDGAN has achieved phenomenal~\cite{Hardy_2019} performance relative to standard FLGAN for independent and identically distributed (iid) datasets.

Similar to online streaming data with varying bit distribution, real-world data distributions are hardly ever iid. In a non-iid setting, each client's dataset may have a different distribution, significantly increasing training difficulty and mode collapse vulnerability. 

Some works~\cite{perfedGan, IFL-GAN} have presented  unique architectures to solve this problem. However, none of these works address the stability problem due to increasing clients. The above discussion raises the following open question:
\textbf{As the number of clients increase, can an architecture maintain high and stable performance for non-iid datasets while avoiding mode collapse?}

We propose MULTI-FLGAN, an architecture inspired by both MDGAN and MGAN. MULTI-FLGAN extends MGAN to a distributed setting to produce an architecture with high and stable performance for non-iid distributions. 
\vspace{0.5em}

\textbf{Contributions}: This paper's contributions are summarized as
\begin{itemize}

 \item  To propose MULTI-FLGAN, a novel architecture that can produce diverse high-quality images with faster convergence than the baseline FLGAN.
 \item  To compare the performance of MULTI-FLGAN against baseline FLGAN and similar competitors over multiple clients using Inception(IS) and Frechet inception scores(FID).
 \item  To compare the learning performance of MULTI-FLGAN against baseline FLGAN and similar competitors on fixed number of clients, i.e. how well MULTI-FLGAN performs over iterations compared to other competitors on IS score.
 \item  To highlight privacy risks and relevant attacks on MULTI-FLGAN. Specifically, Inference attacks and Model poisoning attacks.

\end{itemize}

\section{Preliminaries}

\subsection{Classical GANs}
A classical GAN, as proposed by Ian Goodfellow et al. ~\cite{ClassicalGan} consists of two neural networks: a generator \textit{G} and a discriminator \textit{D}. Their objective can be described by a min-max game, where the discriminator tries to minimize the probability of classifying a fake sample as real while the generator tries to fool the discriminator by producing data similar to the training set.


$\bullet$ {\bf Discriminator Learning Phase:}
The generator takes in a random noise signal \(V\) and generates data \(d_f\)  in the same format as the training data (e.g. 28x28 with 1 color channel as in MNIST dataset). The data \(d_f\) is then fed into the discriminator along with real data \(d_r\). The discriminator acts as a  classifier, classifying whether a sample is fake or real. The classification loss is used to train the weights of the discriminator through back-propagation.

$\bullet$ {\bf Generator Learning Phase:}
The generator learns from the loss of the discriminator. It does so by using the Kullback-Leibler Divergence (KL)~\cite{Joyce2011} loss function. KL quantifies how much the generated distribution differs from the actual distribution. The generator can learn from the discriminator and update its weights by employing this loss function.

\subsection{FLGAN}
FLGAN~\cite{FLGAN} extends classical GAN to a distributed setting. Suppose we have the following setup with N clients:
Each client \(i\) is equipped with a private dataset \(d_i\), a generator, and a discriminator. The main server has a global model \(w\), i.e a global generator and discriminator. A global iteration of FLGAN's learning algorithm is as follows:
\begin{itemize}
    \item Each client trains their local models \(w_i\) and sends the updated weights to the server.
    \item The server aggregates these weights using fedAvg\cite{fedAvg}.
    \item In the next iteration, the server sends the updated global model \(w\) to the clients.
    \item After \(e\) epochs, the main server will have a trained global discriminator and generator. 
\end{itemize}

\section{Problem Formulation}
Suppose N clients wish to learn a global model  \(W \) and agree on a common protocol. Each client  \(i \) has a private dataset \( d_i\) distributed in a non-iid fashion\footnote{This is further explained under Experiment Setup}. This highlights that all clients will have an uneven sample distribution. For instance, client \(A\) may have 100 samples of label \(c\) and only 10 samples of label \(d\), while client \(B\) may possess 10 samples of label \(c\) but 200 samples of label \(d\) . 

The clients agree beforehand on the number of discriminators \( X\) and generators \(Y \) to use. The problem is to efficiently learn the global model \(W\) by minimizing the error on each local model  \(w_i\) using dataset  \(d_i\).

We seek to develop an architecture to solve this problem, with the following characteristics:
\begin{itemize}
    \item Robust: The architecture should maintain a high IS score even in the case of uneven labels for each client. Moreover,  the algorithm should perform even when limited training samples are available, i.e. 2000 to 5000 samples. 
   
     \item Stable: The architecture should perform in a consistent fashion. The IS score of the generated images should not fluctuate greatly when increasing or decreasing the number of clients. 
 
     \item Performant: The architecture should be able to  generate high quality and diverse images. 
 
\end{itemize}

\begin{figure}[h!]
\caption{High Level Architecture}
\includegraphics[width=3.4in]{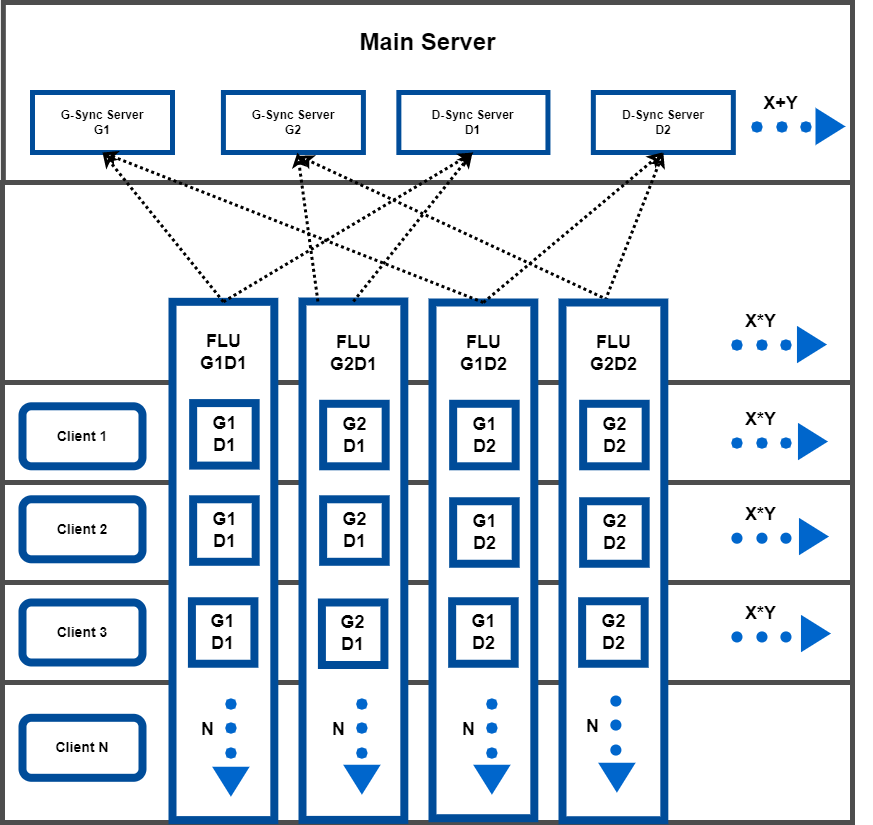}
\centering
\label{overview}
\end{figure}

\section{The MULTI-FLGAN Architecture}

\,\,\,\,\,\textbf{Rationale:} 
Our architecture is mainly inspired by MGAN and MDGAN. By using multiple generators instead of one generator, MGAN successfully avoided the problem of mode collapse while achieving phenomenal performance. Similarly,  MDGAN achieved a comparable performance by having a generator compete against multiple discriminators for a lower computation cost per client.  As shown in Figure~\ref{overview}, MULTI-FLGAN integrates these approaches by introducing an all vs all game with multiple discriminators and generators for each client.

\subsection{Components of architecture} 
\,\,\,\,\,\textit{{Federated Learning Unit(FLU)} }- An FLU contains a cluster of N identical GANs, keeping track of their average generator and discriminator weights. For instance, an FLU \(G2D1\) contains a cluster of \(N\) identical GANs with generator id  \(2\) and discriminator id \(1\).

\textit{{Generator Sync server(G-sync)}} - A G-sync server aggregates the generator weights from its FLUs and stores the resulting weight. For instance, G-sync server G2 aggregates the generator weights of ids \(G2D1\) and \(G2D2\).  

\textit{{Discriminator Sync server(D-sync)}} - Similarly, a D-sync server aggregates the discriminators' weights  from its FLUs and stores the resulting weight. 

\textit{{Sync server(s)}} - A general component referring to both D-sync and G-sync servers.

\textit{{Main server(m)} }- This component is responsible for both allocating and initiating connections between FLUs and Sync servers. 

\vspace{0.5em}
\textbf{Protocol:}
MULTI-FLGAN  follows the following protocol:
\begin{itemize}
 \item \textbf{Step 1: Protocol Initiation}\newline
 The client initiates the protocol by sharing parameters \(X\) and \(Y\) with the server.

\item \textbf{Step 2: Sync Server Allocation}. \newline
 Main server \(m\) allocates \(X\) D-sync servers and  \(Y\) G-sync servers. For example,  in Figure~\ref{overview}, \(m\) allocates 4 Sync servers in total. 

\item \textbf{Step 3: FLU Allocation}. \newline
The main server allocates \textit{X*Y} FLUs. Each Sync server then connects to its respective FLU. In Figure~\ref{overview},  G-sync server \(G1\) connects to FLU ids \(G1D1\), \(G1D2\), and G-sync server \(G2\) connects to ids \(G2D1\), \(G2D2\). Respectively, D-sync server \(D1\) connects to ids  \(G1D1\), \(G2D1\) while \(D2\)  connects to ids  \(G1D2\) and \(G2D2\).

\item \textbf{Step 4: Client Distribution}. \newline
Each FLU will create a partition for each client by replicating identical GANs. For instance, FLU \(G2D1\) creates identical replicas of GANs with generator  \(G2\)  and discriminator \(D1\) for each client. 

\end{itemize}

\begin{table}[h!]
  \caption{Summary of notations}
  \setlength{\tabcolsep}{0.7\tabcolsep}
  \centering
  \begin{tabular}{ *{5}{c} }
    \toprule
    \textbf{Notation} & \textbf{Description} &  \\
    \midrule
    \(D\) & Global Dataset \\
    \(W\) & Global Model \\
   \(d_i\) & Training dataset of client \(i\)      \\
    \(e\) & epochs\\
    \(X\) & Number of discriminators \\
    \(Y\) & Number of generators  \\
     \(G\) & Generator \\
    \(D\) & Discriminator \\
    \(fl\) & Set of all FLUs\\
    \(s\) & Set of all Sync servers\\
    \(s_j\) &  Sync Server j\\
    \(w_i\) & Local model of client \(i\)\\
    \(w_j\) & Local model of \(s_j\) \\
    \(w_a\) & Local model of FLU a (\(fl_a\))\\

    \bottomrule
  \end{tabular}
\end{table}

\subsection{Learning Algorithm}
The goal of our algorithm is to minimize the empirical loss  \(F_w\) of the global model \(W\) on dataset \(D\),  with global learning rate \(\alpha\) and  \(e\) epochs. However, in a distributed setting, each client will solve the optimization problem of \( min_wf(d_i, \left\{ {{w_i}_1} ,..., {{w_i}_k}  \right\}\)) , where \(f\) is the empirical loss of local models \(\left\{ {{w_i}_1} ,..., {{w_i}_k}  \right\}\) under the client \(i\) and  \(k\) is bounded by \(X*Y\).  Our algorithm takes the following steps to solve this objective:

\begin{itemize}
\item \textbf{Step 1: Synchronization of global model  with FLU}\newline
Each Sync server \(s_j\) has a Sync model \(w_j\). \(s_j\) sends \(w_j\) to its respective FLU \(fl_a\), which updates its model \(w_a\) with the incoming model \(w_j\). 

\item \textbf{Step 2: Training local models}\newline
Each client \(i\) trains local models \(\left\{ {{w_i}_1},...,{{w_i}_k}  \right\}\) with learning rate \(\alpha*N\) and a randomly selected and permuted batch of training dataset  \(D_i\).To force the models to converge, we increased the learning rate
proportionally to the number of clients. Additionally, we allow for diversity by
training each model with a randomly selected and permuted batch. This enables GANs to
generalize over uneven distributions.

\item \textbf{Step 3: Update FLU through aggregation}\newline
Upon training, each FLU aggregates the weights of its
generators and discriminators using fedAvg~\cite{fedAvg}. Then the FLUs send the updated models to their respective Sync
servers.

\item \textbf{Step 4: Update Sync servers through aggregation}\newline
Each G-sync  and D-sync server average the generator and discriminator weights of  connected FLUs, respectively. Then, they update their Sync model \(w_j\) with these new weights.

\item \textbf{Step 5: Termination }\newline
After \(e\) epochs, the main server chooses the best generator out of
all G-sync server models using a generic metric such as inception score.
\end{itemize}

\begin{algorithm}[h!]
\SetAlgoLined
\SetAlgoNoLine%
\begin{algorithmic}
  \\

\Procedure{Sync}{$X$, $Y$, $fl$, $s$ }
  \State   \For{j = 1, 2 ...  X+Y}{
      \For{a = 1, 2 ... X*Y }{
            \If{\Call{isConnected}{$s_j$, $fl_a$}} {
               \State  \Call{Send}{$s_j$, $fl_a$, $w_j$}
                \State Update ($w_a$, $w_j$)
            }
        }
}
\EndProcedure

\Procedure{TrainFLU}{$N$, $X$, $Y$ }
   \State  \For{i = 1, 2 ... N}{
          \For{a = 1, 2 ... X*Y }{
          \State $batch_i = $ \Call{GetRandom}{$D_i$}
          \State  \Call{TrainLocal}{$w_{i_a}$,$batch_i$,  $\alpha*N$, e}
     }
}
\EndProcedure
 
\Procedure{UpdateFLU}{$N$, $X$, $Y$,  $fl$}
      \State  \For{a = 1, 2 ... X*Y }{
         \State $sum_{g}$ = 0
         \State $sum_{d}$ = 0
       
         \State  \For{i = 1, 2 ... N}{
              \State $sum_{gen}$  =    $sum_{g}$ + \Call{getGWeight}{$fl_{a_{i}}$}
                 \State $sum_{disc}$  =    $sum_{d}$ + \Call{getDWeight}{$fl_{a_{i}}$}
               
         }
         \State  $fl_a$.\Call{setGWeight}{$\frac{sum_{g}}{N}$}
         \State  $fl_a$.\Call{setDWeight}{$\frac{sum_{d}}{N}$}
   
         }
   \EndProcedure

  \Procedure{UpdateSyncServer}{$N$, $X$, $Y$,  $fl$} 
       \State  \For{j = 1, 2 ...  X+Y}{
         \State $s_{g}$ = 0
         \State $s_{d}$ = 0
          \State \For{a = 1, 2 ... X*Y }{
                \If{\Call{isConnected}{$s_j$, $fl_a$}} {
                  \State $s_{g}$ = $s_{g}$ + $fl_a.\Call{getGWeight}{ }$
                  \State $s_{d}$ = $s_{d}$ +$fl_a.\Call{getDWeight}{ }$
                }
            }
        \State  $s_j$.\Call{setGWeight}{$\frac{sum_{g}}{X}$}
         \State  $s_j$.\Call{setDWeight}{$\frac{sum_{d}}{Y}$}
        }
 \EndProcedure 
 
\Procedure{Termination}{$N$, $X$, $Y$, $S$} 
         \State $score$ = 0
          \State $best_w$ = None
            \State \For{j = 1, 2 ...  X+Y}{
               \If{$score > $ \Call{getScore}{$s_j$}} {
           \State $score$ = \Call{getScore}{$s_j$}\
                \State $best_w$ = $s_j$
               }
        }
        
       \State \Return \(best_w\) 
        
 \EndProcedure

\end{algorithmic}      
 
 \caption{The Multi-FLGAN Algorithm}
\end{algorithm}
 \vspace{1em}

\subsection{Client Model Learning Procedure} All client models are trained in batches of 64 samples. From a random noise vector \(V\) in each iteration, the generator generates a \(D^{f}\) batch with the same dimensions as a real sample. This is sent along with real  batch  \(D^{r}\) to the discriminator. The discriminator uses the loss of miss-classifying fake samples to update its weights. Similarly, the generator uses the KL \cite{Joyce2011} loss function to learn the true training sample distribution from the discriminator.
\vspace{1em}



\section{Evaluation}

\subsection{Experimental Setup} 
 \textbf{Computational Setup:}
Recently, there has been a trend in adapting serverless \cite{Hardy_2019} frameworks(e.g. gossip protocol introduced by Dynamo back in 2007). However, the case for deep learning and machine learning systems is not yet sufficient. Distributed deep learning systems use massive amounts of data that use data-intensive operations such as back and forward propagation, making it necessary to operate in a parallel environment. We extended the implementation\footnote{https://github.com/bbondd/DistributedGAN} of Distibuted GAN using Ray to emulate a parallel environment for distributed learning.  Ray \cite{Ray} was mainly used due to its superior performance benchmarks in distributed machine learning instead of frameworks like Dask~\cite{Dask}.
 
Ray's distributed framework uses a central Redis server to simulate a head node responsible for communication and aggregation between different worker nodes. We simulated the main server as the head node and each Sync server as a worker node. We also assigned worker nodes for each FLU to simulate multiple FLUs under a sync server. In total, we used 8 worker nodes - 4 for Sync servers and 4 for FLUs.

Moreover, each GAN was given its own thread for each FLU worker node. This allowed for parallel training across all FLUs over multiple nodes. Of course, such an elaborate setup introduces latency and bandwidth issues. Nevertheless, such an environment was necessary to mimic real life use cases.

 \textbf{Datasets:}
 We experimented with two classical datasets used for machine learning: MNIST~\cite{deng2012mnist} and Fashion MNIST (FMNIST)~\cite{xiao2017fashionmnist}. The MNIST and FMNIST datasets are comprised of 60,000 28$\times$28-pixel grayscale image samples of handwritten digits, and clothing, respectively.

 \textbf{Sampling Process:}
To distribute \(k\) samples in a non-iid fashion from a global dataset \(D\), we took a unique approach.  We  randomly selected 5000 samples from \(D\)  as the training set \(t\),  to emulate the lack of data and  decrease the training time. Next, we reserved a random fraction \(F_i\) of \(t\), using  Mersenne Twister~\cite{Twister}, for each client. Each  \(F_i\) was then divided into batches of 64 samples and assigned to their respective client.

 \textbf{Hardware:}
Our experiments were made using a TensorFlow backend with 4 NVIDA Tesla v100 GPUs and 100 CPU cores made available by High Performance Computing Platform~\cite{DHPC2022}. The head node was given 44 cores and each worker node was assigned a partition of available GPUs. 

 \textbf{GANs Architectures:}
 In our experiments, we used a traditional type of GAN named DCGAN~\cite{Rathford}
 . The generator is comprised of six transposed convolution layers of 128, 64, 32, and 1 with kernels of size 5 x 5. On the other hand, the discriminator uses four transposed convolution layers of 32, 64, 128 and 256 with kernels of size 5 x 5 and a fully connected dense layer. 
 Typically, a dropout layer is also included in both of these neural networks. Instead, we opted for batch normalization, as it helps quicken the learning process. 
 
  \textbf{Hyperparameters:}
 Since the dropout layer was removed, we introduced a decay factor to prevent over-fitting. In order to fine-tune the hyperparameters of a standalone DCGAN, we performed grid search, varying the decay factor and learning rate. We found that the optimal decay factor was \(1.5e^{-8}\) with a fixed learning rate of \(2e^{-4}\).
  
  \textbf{Competing Approaches:} We tested our architecture against traditional FLGAN and  AFLGAN, a variant of MDGAN that aggregates only the generator weights while leaving the discriminator weights unchanged across iterations. We simulated AFLGAN and FLGAN using the same Ray setup and hardware. The main server was assigned its own head node, while each client was assigned one worker node.

\textbf{Metrics:}
Evaluating GANs is often difficult as an objective metric is needed to capture image diversity and quality. We must ask, "Do images look like a specific object" and "Is a wide range of objects generated?"

A popular metric, Inception Score (IS)~\cite{salimams} addresses quality by considering how strongly an image is classified as one class over others. Similarly, it considers diversity by examining the marginal probability distribution of the generated images. Typically, a low inception score indicates low quality and uniform image, while a high inception score implies diverse and high-quality images.

However, the inception score does not reveal how far off the generated images are from the real images. Therefore we also use Frechet Inception Distance~\cite{FID} or FID for short, as proposed by Heusel et al. FID calculates the distance between feature vectors of the original set of images and the generated set of images. In contrast with the IS score, a low FID score is preferred, as it implies that the difference between the distributions is small.

\begin{table}[h!]
  \caption{Experiment Parameters}
  \setlength{\tabcolsep}{0.7\tabcolsep}
  \centering
  \begin{tabular}{ *{6}{c} }
    \toprule
    \textbf{Experiment}  & \textbf{IV} & \textbf{DV}
    & \textbf{Samples} & \textbf{LearningRate}  \\
    \midrule
    Type \( 1\) & Clients & IS/FID  & 5000 & 0.0002 \\
    Type \( 2\) & Iterations  & IS & 5000 & 0.0002\\
    \bottomrule
  \end{tabular}
\end{table}

\textbf{Experiments:}
We conducted two kinds of experiments. First, we tested how well each architecture performs by varying the number of  clients $N \in [2,3,5,10,20]$ on both datasets(FMNIST and MNIST).  Each experiment was performed for 100 epochs with noise vector \(V\) of dimension 100. IS and FID scores were computed every 10 epochs. Note that FID scores were computed using the test dataset dimensions: 10000 samples.

Second, we use IS score to compare how well different architectures learn through iterations $e \in [0, 10, 20, 30, 40, 50, 60, 70, 80, 90]$  for fixed clients $N \in [2, 3, 5, 10,20]$ on both datasets(FMNIST and MNIST).

The experiments used MULTI-FLGAN with only two discriminators and generators. We opted for 2 discriminators and generators to demonstrate that even with minimal parameters, our architecture is far more performant than other alternatives. While this may not be the ideal amount of discriminators and generators, it suffices for the purposes of this study. Testing precise head-to-head comparisons between different discriminators and generators is left to future work.

\subsection{Experimental Results}
\begin{figure*}[ht] 
\centering
\caption{ Experiment Scores}
\includegraphics[width=1\linewidth]{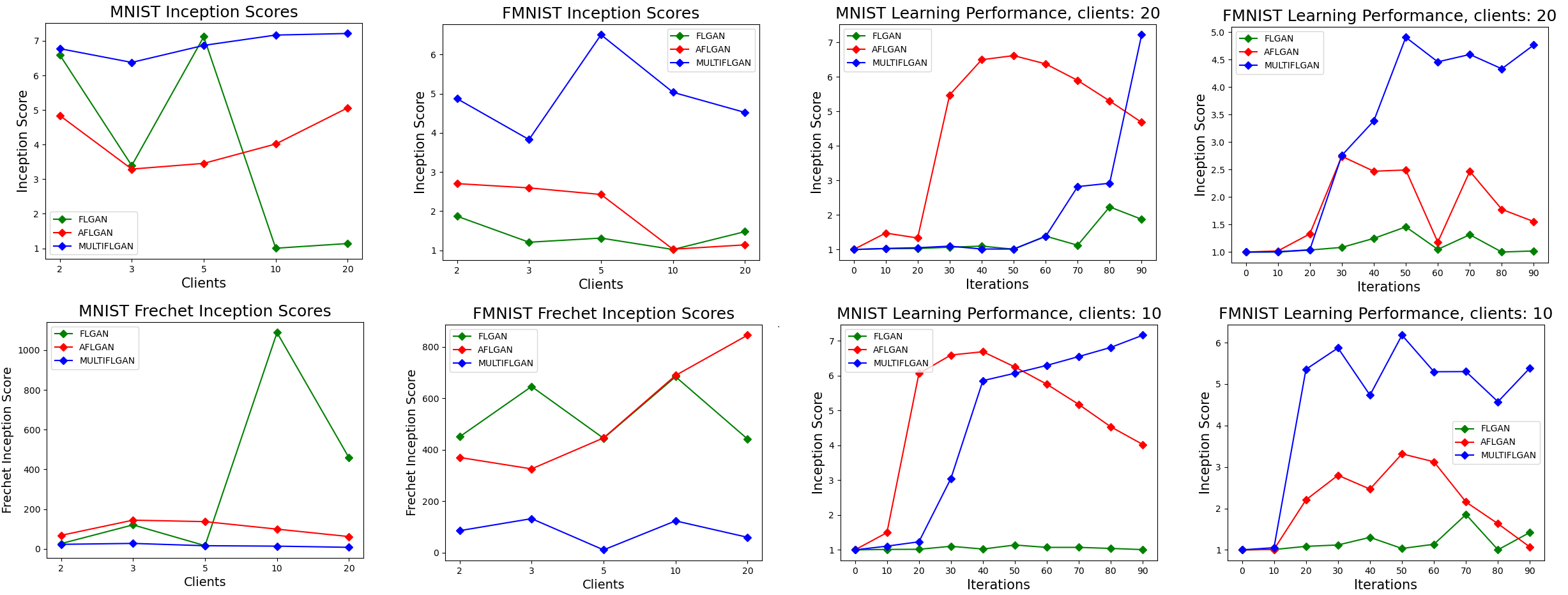}
\label{scores}
\end{figure*}

\begin{figure*}[ht] 
\subsection{Learning Performance Graphs}
\centering
\caption{Learning performance for\textbf{ \#Clients: 2, 3, 5}}
\vspace*{3mm}
\includegraphics[width=1.0\linewidth]{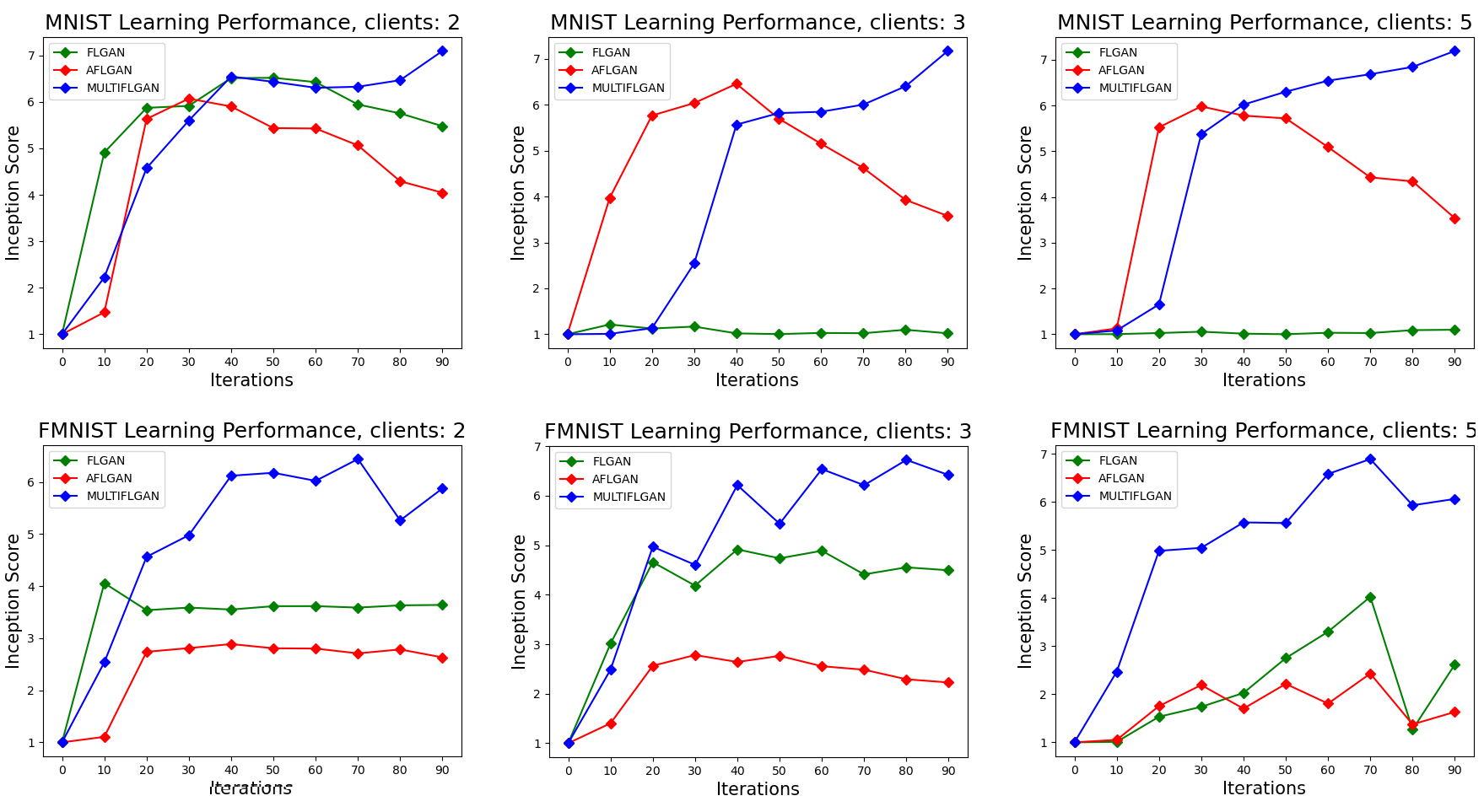}
\label{score2}
\end{figure*}

Tables~\ref{fmnistis} and~\ref{mnistis} report the average, min and max FID and IS scores of over 20 clients for experiments testing clients' performance. Additionally, Figure~\ref{scores} and~\ref{score2} depicts graphs\footnote{Note that learning performance graphs for 2, 3 and 5 clients are presented in Figure~\ref{score2}  } of the client and learning performance experiments.

    \begin{table}[h!]
  \caption{Client Performance of FMNIST(IS/FID)}
  \setlength{\tabcolsep}{0.7\tabcolsep}
  \centering
  \begin{tabular}{ *{6}{c} }
    \toprule
    \textbf{Heuristic}  & \textbf{AFLGAN}
    & \textbf{FLGAN}     & \textbf{MULTI-FLGAN} \\
    \midrule
    Min & 
    1.00 / 328.10   & 1.00 / 442.41 & 3.82 / 11.60 \\
   Max & 2.70 / 847.00 & 1.89 / 646.39  & 6.48 / 129.35 \\
     Average & 2.00 / 535.30  & 1.37 / 495.75 & 4.95 / 82.33\\
    \bottomrule
  \end{tabular}
  \label{fmnistis}
\end{table}

 \textit{1) \textbf{Competitor score for FMNIST:}}
With an average IS score of 1.37 and an FID score of 495.75, we observe that FLGAN performs the worst relative to its competitors. It is also apparent from the FMNIST dataset inception scores shown in Figure~\ref{scores} that FLGAN struggles to generate high-quality data, having only achieved a maximum IS score of 1.89 for two clients. The learning performance of the FMNIST dataset is also unimpressive. For 2 to 3 clients, it reaches an average IS score of 4.5. For 5 clients, performance drops further to 2, and for 10 to 20 clients, hardly any learning occurs. This poor performance is further reflected by the generated 
images, which gradually become undecipherable with increasing clients.

On the other hand, AFLGAN performs relatively well compared to FLGAN for 2, 3, and 5 clients but drops to a minimum inception score of 1 for 10 and 20 clients, resulting in an average inception score of 2.00 and a FID score of 535.00. This sudden drop implies that AFLGAN cannot maintain its performance over increasing clients.

 Interestingly, the learning performance graphs of FMNIST for 10 and 20 clients show that AFLGAN peaks at 30 - 40 iterations, even matching MULTI-FLGAN performance, before suddenly decreasing in performance. 

We believe that this behavior is a consequence of uneven training samples. For instance, if generator \(G1\)  trains on 200 samples of label \(A\) while generator \(G2\) trains on only 10 samples of label \(A\), their weights will differ considerably. This leads to not only training instability, but also failure to converge, as evident in Figure~\ref{scores}. Of course, it is unrealistic to expect convergence from only 100 iterations. Having said that, the parabolic behavior is proof of AFLGAN's inability to converge.

In contrast, MULTI-FLGAN tackles uneven distributions by having multiple generators compete against multiple discriminators. This results in more robust scores, as depicted in Table 3. While the fluctuation between 2, 3 and 5 clients shows that uneven label distributions still influence our architecture, it still maintains an average IS score of 4.95 and a 82.33 FID score, outperforming the other competitors. Additionally, the IS and FID graphs for FMNIST shown in Figure~\ref{scores} make clear that our architecture is performant even with 10 and 20 clients, achieving four times AFLGAN and FLGAN scores.

  \begin{table}[H]
  \caption{MNIST(IS/FID)}
  \setlength{\tabcolsep}{0.7\tabcolsep}
  \centering
  \begin{tabular}{ *{4}{c} }
    \toprule
    \textbf{Heuristic}  & \textbf{AFLGAN}
    & \textbf{FLGAN}     & \textbf{MULTI-FLGAN} \\
    \midrule
    Min & 
    3.30/ 60.70
 & 1.00/ 15.46& 6.39/ 8.07\\
   Max & 5.10/ 141.90 & 7.10/ 1087.00 & 7.15/ 26.50\\
     Average & 4.10/ 101.20  & 3.81/ 341.34 & 6.84/ 17.10\\
    \bottomrule
  \end{tabular}
  \label{mnistis}
\end{table}
  
  \textit{2) \textbf{Competitor score for MNIST:} }
 All architectures performed much better on MNIST than FMNIST, as seen on table 4, since it is much easier to learn the distribution of numbers than complex fashion designs with only 5000 samples and 100 epochs.
 
 Interestingly, FLGAN experienced a drop in IS score from 7 to 1 between 5 and 10  clients. Inspecting the generated images 
 , it indicates that FLGAN experienced mode collapse, resulting in an average IS score of 3.81 and FID score of 341.34. In this case, the mode collapse happened at iteration 50 where FLGAN exclusively generated the number 3. Subsequent iterations show that FLGAN was unable to generalize over other numbers, leading to a meaningless final outcome.

 In contrast, AFLGAN performed exceedingly well. It  sustained an average inception score of 4.10 and an FID score of 101.20. Although it exceeds the learning performance of MULTI-FLGAN in the first few iterations for clients 10 and 20, AFLGAN performance gradually decreases over subsequent iterations. This parabolic behavior reconfirms its inability to converge. 
 
MULTI-FLGAN, on the other hand, maintained the same performance using MNIST as FMNIST, resulting in an average IS score of 6.84 and an FID score of 17.10 without signs of mode collapse.  Interestingly, the learning performance graphs for 10 and 20 clients show a sudden jump in improvement for both MNIST and FMNIST. For MNIST, the jump occurs at 50 iterations for 20 clients, while for FMNIST, the jump occurs at 20 iterations for 20 clients. This may be due to a lack of encountered samples in the first 20 or 50 iterations. However, once  MULTI-FLGAN sees sufficient samples, its performance increases exponentially. 
 
It is also interesting to compare the stability of MULTI-FLGAN with other competitors. We quantify  \textit{stability} as the difference between the max and min of FID scores. Intuitively, the distance between generated and actual images should not drastically fluctuate between clients. 


 \begin{table}[H]
  \caption{Stability(MNIST/FMNIST)}
  \setlength{\tabcolsep}{0.2\tabcolsep}
  \centering
  \begin{tabular}{ *{4}{c} }
    \toprule
    \textbf{Heuristic}  & \textbf{AFLGAN}
    & \textbf{FLGAN}     & \textbf{MULTI-FLGAN} \\
    \midrule
    Stability & 81.2/518.9 & 1071.9/203.9 & 18.43/117.8\\
    \bottomrule
  \end{tabular}
\end{table}

Table 5 shows that MULTI-FLGAN is at least 30 times as stable as AFLGAN and 58 times times as stable as FLGAN. Trained on FMNIST, FLGAN's stability is deceptively close to that of MULTI-FLGAN. Note, however, that FLGAN's FID score is far higher relative to MULTI-FLGAN, indicating that the generated images vary significantly from the actual distribution. 

\vspace{0.5em}
\textit{3)\textbf{ Overall Performance:}}
FLGAN performs poorly on both datasets, struggling to converge on higher number of clients. AFL-GAN produces decent results for the first 5 clients but then experiences a sudden drop when generalizing over 10 or more clients. On the other hand, MULTI-FLGAN has outperformed both architectures on both datasets. The graphs and the generated images show that MUTLI-FLGAN can both generalize over uneven samples and produce high-quality images with stable IS and FID scores over increasing clients. \vspace{0.5em}

  \textit{4)\textbf{Trends:}}
It is hard to predict any architecture's learning performance trend for 50 or 100 clients. However, seeing that the learning performance of AFLGAN has already peaked within 100 iterations for 10 and 20 clients, it is safe to assume that it will only perform worse on a higher number of clients. Since FLGAN  reached a minimum IS score of 1 on both data sets for ten clients, it cannot be reasonably expected to learn for a higher number of clients. However, we cannot rule out the possibility of FLGAN realizing a jump similar to MULTI-FLGAN when given enough samples. 

The learning performance graphs of FMNIST show that MULTI-FLGAN is stabilizing around iteration 70, while on MNIST, it is still increasing beyond 90 iterations. Regardless, the IS and FID scores lead us to believe that even for clients above 50, MULTI-FLGAN would be able to maintain the average inception score without significant fluctuations.



\section{Discussions \& Improvement}
MULTI-FLGAN has shown promise in terms of performance, robustness, and stability.
Nonetheless, we must still consider other aspects such as cost, scaling, and security prior to deployment. 

\textbf{Complexity:}
One of the main concerns is the architecture's suitability for the storage and computation limitations of small devices such as mobiles. To aid the discussion on computation and space complexity, we will introduce new notations in addition to Table 1.  Let  \(o\) be the object size(e.g image size in MB), \(b\) batch size and \(D_{i_{o}}\), number of objects in local dataset \(D_i\) of client \(i\). 

The computation complexity and space complexity of FL-GAN and MULTI-FLGAN have been summarized in Table~\ref{complexity}.

\begin{table}[h!]
  \caption{Complexity}
  \setlength{\tabcolsep}{0.7\tabcolsep}
  \centering
  \begin{tabular}{ *{3}{c} }
    \toprule
    \textbf{} & \textbf{FLGAN}    & \textbf{MULTI-FLGAN}\\
    \midrule
    O(C) & \(O(eb\sum_{i=1}^{N}\frac{\left | w_i \right |}{D_{i_{o}}})\) & O(\(eb(\sum_{i=1}^{N}\sum_{j=1}^{XY} \frac{\left | w_{i{_j}}\right |}{D_{i_{o}}} )\) \\[2ex]
  O(S) & \(O(\sum_{i=1}^{N} \left | w_i  \right |)\) & \( \sum_{i=1}^{N}\sum_{j=1}^{XY} \left |w_{i_{j}}  \right |)\)\\
   
    \bottomrule
  \end{tabular}
  \label{complexity}
\end{table}

Admittedly, the computation complexity of MULTI-FLGAN is much more expensive than traditional FL-GAN as it depends on both \(N\) and \(XY\). 
Furthermore, in practice, there is the additional cost of communication between the main server, Sync servers and  FLUs. Table~\ref{time} presents the computation time of different architectures used in experiment type 1.



\begin{table}[h!]
\centering
\caption{Time taken for Experiment Type 1}
\setlength{\tabcolsep}{0.7\tabcolsep}

\begin{tabular}{@{}llllll@{}}
\toprule
Architecture & \multicolumn{5}{c}{Time Taken}                                   \\ \midrule
             & \multicolumn{5}{c}{Clients}                                      \\
             & \textbf{2} & \textbf{3} & \textbf{5} & \textbf{10} & \textbf{20} \\ \cmidrule(l){2-6} 
FLGAN        & 0.1        & 0.2        & 0.6       & 2.5        & 4.5         \\
AFLGAN       & 0.1        & 0.2        & 0.7        & 2.4         & 4.5        \\
MUTLI-FLGAN  & 0.2        & 0.4        & 6.0        & 12.0        & 23.0        \\ \bottomrule
\end{tabular}%
\label{time}
\end{table}

We  notice that MULTI-FLGAN takes almost 6 times as long as FLGAN and AFLGAN. However, there are several adaptations we can make to improve this computational complexity. One possible way is to adapt a similar technique to MDGAN, where the number of FLUs are significantly reduced.

The strategy is to divide each combination of generators and discriminators among the clients avoiding the  need for replication. Consider the case with two generators, three discriminators, and three clients. Client \(A\) will train \(G1D1\) and \(G1D2\), Client \(B\) - \(G2D1\) and \(G2D3\), and Client \(C\) - \(G2D2\) and \(G1D3\) reducing the number of FLUs needed to two. Eventhough, there is a significant improvement in the computation complexity, each client now trains a fraction of the models in the former architecture. To compensate for the lack of models,  we will introduce a  peer-to-peer mechanism  where each discriminator and generator is swapped with a random client every two epochs . 

The adapted strategy results in a drastic decrease in computation cost while preserving the same notion of an All vs ALL game. A possible drawback of using this adapted approach is connectivity. If a client disconnects during training, the algorithm loses unique pairs of generators and discriminators hindering the performance of the global model.

As seen from the space complexity of table 4, our algorithm uses much more memory than baseline FLGAN. However, we assume clients that adopt our protocol would meet the computational and spatial requirements. Besides most modern smart devices can easily handle 5 to 10 MB of data(Size of a model). Therefore we believe that our architecture is still feasible for a rational choice of  \(X \) and  \(Y \).

\textbf{Scalability:}
The second aspect to consider is whether our architecture is readily scalable. Currently, our architecture can accommodate any amount of clients with arbitrary parameters X and Y. Any client may participate in the training  as long as they join at Step 1 of our learning procedure using  the same X and Y parameters as other clients. If not, the client will be asked to wait until it reaches step 1 of our learning algorithm again for synchronization. In this sense, our architecture is scalable on demand. 

However, our architecture does not currently support clients with different X and Y parameters.  But by clustering FLUs based on X and Y, we can allow multiple clients with different parameters to train simultaneously.  

\textbf{Fault Tolerance:}
Our architecture is also resilient to node failures. If one FLU fails, then the respective Sync server will simply continue aggregating models from other connected FLUs. Likewise, if a Sync server fails the training would still continue with  other available Sync servers. The same principle holds for a client who disconnects form the main server.

\textbf{Security:}
Federated learning is vulnerable to different types of attacks. We will mainly consider Inference and Model poisoning attacks.

We  consider a similar attack model to that of FLTrust~\cite{FLTRUST}. More specifically, an attacker manipulates a minority of malicious clients, which can be fake clients injected by the attacker or genuine clients compromised by the attacker. The malicious clients can send arbitrary updates  with the intent of destroying the model or inferring the private data of other clients. We assume that the attacker has full knowledge and access to the following information:
learning rate, models and latent vector of compromised clients. 

\small
\textbf{Inference Attacks:}
\normalsize
In an inference attack, the attacker is interested in inferring the private data of other clients. Our architecture is particularly vulnerable, as each client has access to a partition of all models in FLUs. 

\begin{figure}[htp]
 
    \caption{ Inference attacks}
    \center
    \includegraphics[width=0.5\linewidth]{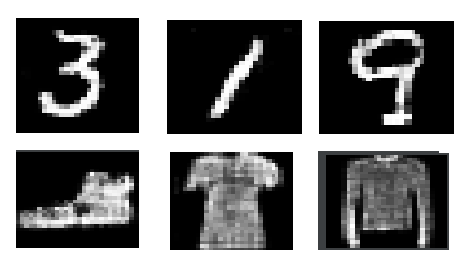}
    \label{inference}
  \end{figure}

We assume the following setting: \(N\) clients train MULTI-FLGAN with 2 discriminators and generators on both MNIST and FMNIST dataset. Each malicious client \(j\) does not actively participate in training  models \(\left\{ {{w_j}_1} ,..., {{w_j}_k}  \right\}\) , instead they return the models as it is.  After \(e\) epochs, the attacker can simply use  any of the compromised clients' generators to produce images from random noise. These generated images can then be easily reconstructed as described by hitaj et al. \cite{inference}.  We have presented the result of such an attack with 15 genuine clients and 5 malicious clients in Figure~\ref{inference}.
\setlength{\abovecaptionskip}{0pt plus 0pt minus 1pt}

From Figure~\ref{inference}, it is clear that the malicious clients could successfully infer private training data through this attack.  Inference attacks can be mitigated using differential privacy. Differential privacy obfuscates the generator weights by adding Gaussian noise to protect the privacy of the training data set. 

Typically, the noise level can have a perturbing effect on the performance of the global model. A lot of research was done to mitigate its affect on the IS score of the generated images. For example, Xie et al.   \cite{Xie}  suggested using DP with carefully designed noise vectors with gradient clipping to minimise the impact on the global model. Similarly, Xinaglong et al.  \cite{AntiGAN} proposed training generators, with a special loss function that obfuscates only the visual features. Both these methods cause little harm to the overall performance of the model. Our architecture could be adapted to use either of these methods by changing the Client learning procedure.

\small
\textbf{Model Poisoning Attacks:}
\normalsize
Another common family of attacks are model poisoning attacks. In this scenario, the attacker wants to destroy or force the global model only to generate images that the attacker is interested in. The only difference from the previous scenario is that the malicious clients update the models with random weights instead of returning the model. Admittedly, we have not conducted any experiments to deduce the effect of model poisoning on our architecture. However, we would like to remark that this is very similar to having uneven distributions for each client. Therefore, we believe that for a minority of clients, our architecture would still be quite resilient. 

However, the averaging operation used by the FLUs is quite sensitive to differences in weight. This operation could be improved by replacing the it with criteria-based aggregation methods. For instance, one could use the inception score as a metric to choose the best model for the next iteration or even other statistical methods such as trim-mean~\cite{trim}, median and Krum~\cite{KRUM}.

\section{Conclusion}
We proposed General Adversarial Networks in the novel context of distributed learning to solve the instability and model collapse problem for non-iid datasets. We drew inspiration from MDGAN and MGAN to develop an all vs all game that allows GANs to easily generalize over uneven labels . Our extensive evaluations of 2 datasets show that our architecture can achieve phenomenal performance and stability with minimal discriminators and generators. In particular, MULTI-FLGAN was able to achieve to maintain the highest average IS score over the 20 clients we experimented with. However, in our research, we tested MULTI-FLGAN only on two grayscale datasets. It would be interesting to see how well MULTI-FLGAN performs on coloured datasets and with different  parameters. We believe this work has introduced a viable solution for the instability problem experienced by federated GANs and hope that raised perspectives will inspire future works.

\section{}
\appendix
\section{Appendix}
\subsection{DCGAN Architecture}
\begin{figure}[h!]
\caption{ Generator}
\includegraphics[width=5cm]{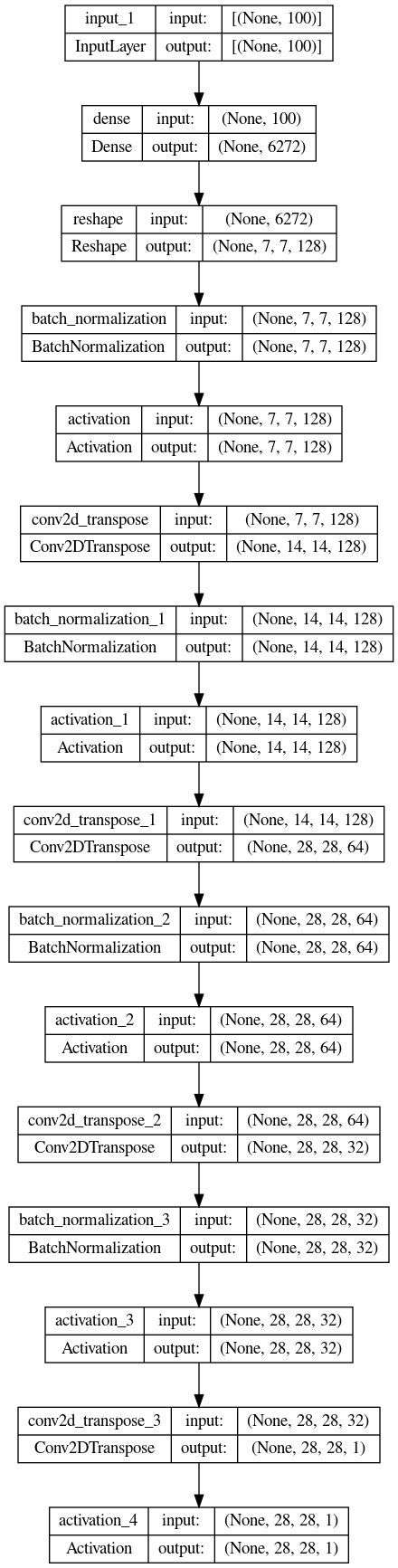}
\centering
\label{generator}
\end{figure}

\begin{figure}[h!]
\vspace{6.5em}
\caption{ Discriminator}
\includegraphics[width=5cm]{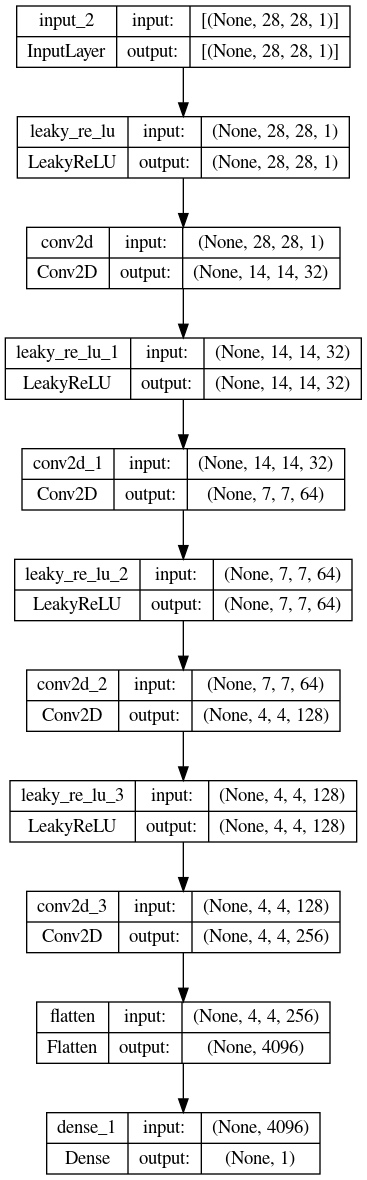}
\centering
\label{discriminator}
\end{figure}

The specific architecture of DCGAN can be seen in the Figure~\ref{generator} and~\ref{discriminator}.

\section*{Acknowledgment}

The authors would like to thank...

\ifCLASSOPTIONcaptionsoff
  \newpage
\fi



%
\printbibliography



%








\end{document}